\documentclass[10pt,twocolumn,letterpaper]{article}

\usepackage[pagenumbers]{cvpr} 

\usepackage[table]{xcolor}

\usepackage{bbm} 

\definecolor{cvprblue}{rgb}{0.21,0.49,0.74}
\usepackage[pagebackref,breaklinks,colorlinks,allcolors=cvprblue]{hyperref}

\def\confName{CVPR}
\def\confYear{2027}

\title{CREST: Exemplar-Free Continual Adaptation of Vision-Language Models \\via Shared Subspace Flow Routing}

\author{
Sriram Mandalika,
Alexandra Schild,
Antonio Rueda-Toicen,
Gerard de Melo
\\[0.5em]
Hasso-Plattner Institute, Germany \\
\texttt{\{sriram.mandalika\}@student.hpi.de}
\\
\texttt{\{aleksandra.kudaeva; antonio.rueda-toicen,gerard.demelo\}@hpi.de}
}

\begin{document}
\maketitle
\begin{abstract}
Large vision-language models such as CLIP achieve strong zero-shot transfer across diverse visual domains, yet adapting them sequentially to new tasks risks overwriting prior knowledge and degrading generalization. This challenge intensifies under class-incremental protocols, where absent task identity causes routing failures to cascade into systematic misclassification, producing a persistent performance gap between task-incremental and class-incremental continual learning. We present \textbf{CREST} (\textbf{C}ontinual \textbf{RE}ctified-flow with \textbf{S}hared \textbf{T}ransport), a unified framework addressing parameter efficiency, routing fidelity, and zero-shot preservation within a single cohesive design. CREST introduces a shared adapter that keeps dominant parameter count fixed across the task sequence, a flow-matching-based router operating in the full frozen feature space to maintain discriminative routing as tasks accumulate, and a confidence-driven bypass gate that reverts the visual encoder toward the frozen backbone on out-of-distribution inputs without exemplar storage. On the MTIL benchmark spanning eleven diverse visual domains, CREST achieves 90.1\% Last accuracy under the task-incremental protocol and 89.5\% under class-incremental evaluation, outperforming the prior state of the art by up to 2.9 points and demonstrating that routing quality rather than adapter capacity is the binding constraint in continual vision-language learning.
\end{abstract}    
\section{Introduction}
\label{sec:intro}

\begin{figure*}[t]
\centering
\includegraphics[width=0.9\linewidth]{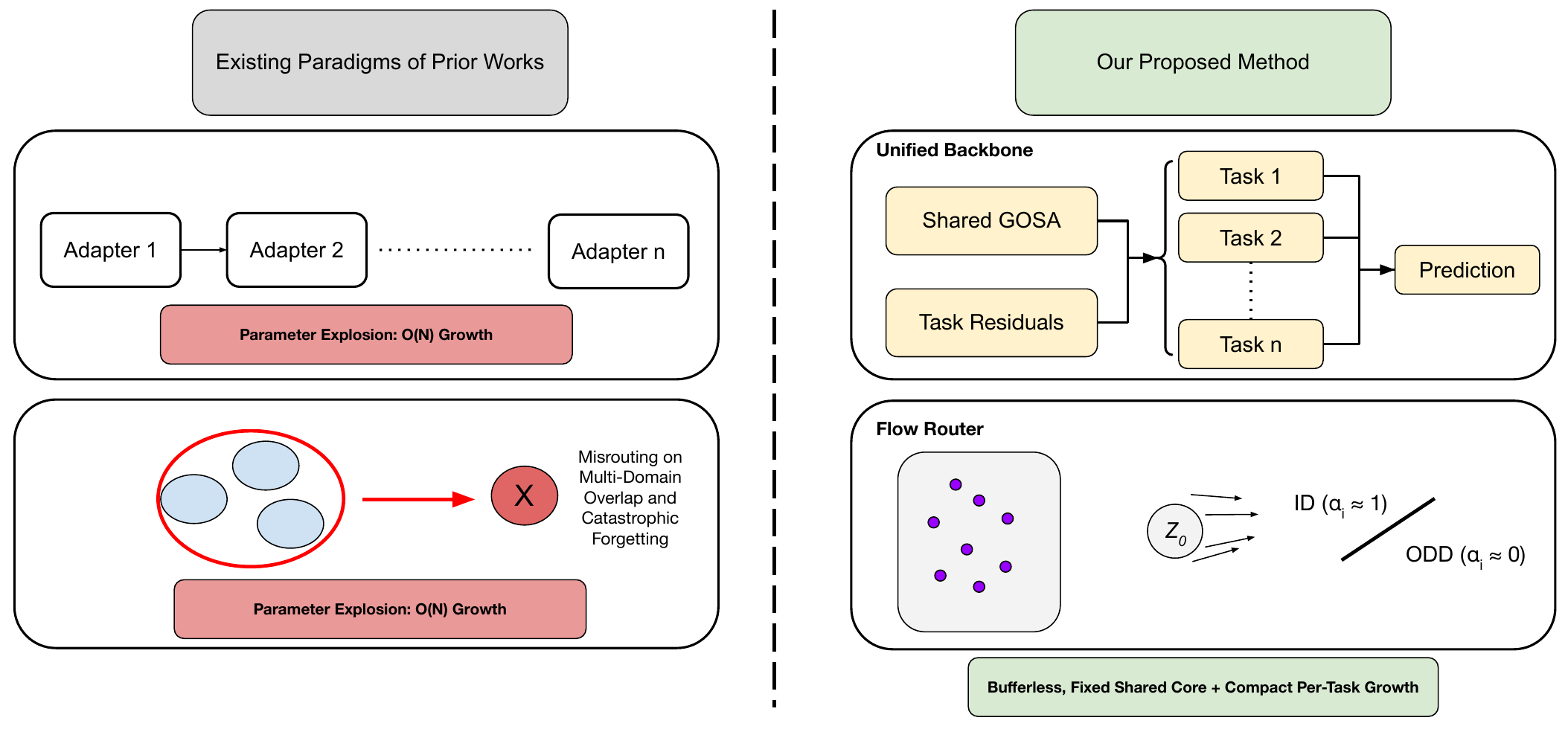}
\vspace{-0.2cm}
\caption{\textbf{Comparison of prior continual PEFT paradigms and CREST.} \textit{(a)} Prior methods allocate independent adapters per task, causing $O(N)$ parameter growth, and route via single-Gaussian distributions that overlap as fine-grained domains accumulate. \textit{(b)} CREST combines a single shared GOSA core with compact per-task residuals and routes queries through a 512-dimensional velocity field trained via conditional flow matching. Conditional velocity-reconstruction error to each task's $K{=}8$ registered centroids selects the active residual at inference, while the bypass gate yields $\alpha \approx 1$ for in-distribution inputs and $\alpha \approx 0$ for out-of-distribution inputs to attenuate visual adapter drift.}
\label{fig:teaser}
\vspace{-0.3cm}
\end{figure*}

Vision-language models trained on internet-scale data have substantially advanced visual recognition, with models such as CLIP~\cite{radford2021learning} achieving strong zero-shot generalization across diverse domains without task-specific supervision. This has motivated considerable interest in adapting
these models to specialized sequential tasks, where new domains arrive one at a time and prior knowledge must be retained without retraining the full backbone.

Parameter-efficient fine-tuning (PEFT) has become the dominant strategy for this setting. Keeping the backbone frozen and learning
lightweight modules such as prompt tokens or low-rank adapters constrains forgetting through structural isolation~\cite{tang2024mind, li2025coleclip, yu2024boosting,
jang2026dpw}. Under the task-incremental learning (TIL) protocol, where task identity is provided at test time, this works well. The class-incremental learning (CIL) protocol removes that assumption: the model must infer the task from the input alone, making routing quality as consequential as adaptation quality.

Existing PEFT continual learning methods share three compounding limitations. First, parameter count grows linearly with every task: dedicated adapters or prompt pools must be allocated per task, limiting long-horizon deployment when the number of future tasks is unknown. Second, routing degrades over time. As fine-grained and semantically adjacent domains accumulate, a single per-task Gaussian cannot represent multimodal intra-task structure, and unimodal densities increasingly fail to discriminate overlapping feature clusters~\cite{zheng2023preventing}. We show that operating directly in the frozen 512-dimensional CLIP space with multi-prototype flow-matching enables reliable task routing even after 11 sequential task updates. Third, task-adapted visual modules that remain fully active on all inputs progressively distort the frozen CLIP feature manifold, suppressing zero-shot transfer on unseen domains.

We propose \textbf{CREST} (\textbf{C}ontinual \textbf{RE}ctified-flow with \textbf{S}hared \textbf{T}ransport), an exemplar-free continual learning system addressing all three limitations in a unified design. Its core adapter, GOSA (Gated Orthogonal Subspace Adaptation), is a shared dual-subspace module injected at every attention and MLP layer across all tasks. Instead of independent per-task adapters, CREST pairs this constant $16.83$\,M shared core with an isolated vision-side rank-$r_p$ residual ($r_p{=}4$) that is frozen after each task, adding only ${\sim}0.59$\,M parameters per task. Without storing exemplars, an exponential moving average (EMA) snapshot of the shared adapter frozen at each task boundary anchors visual features via functional distillation, while centroid-conditioned velocity distillation and cached text-embedding alignment prevent routing and cross-modal drift. For routing, the \textbf{Flow Router} learns a prototype-conditioned velocity field via conditional flow matching on frozen CLIP features and assigns queries to the task whose nearest centroid minimizes conditional velocity-reconstruction error. A residual bypass gate then scales the visual adapter contribution by routing confidence, reverting the image encoder toward frozen CLIP when queries fall far from known task prototypes to preserve zero-shot transfer.

Our contributions are threefold:

\begin{itemize}

    \item \textbf{CREST, a shared-subspace continual learning
    architecture for VLMs.} A shared GOSA core provides
    constant-capacity task-common representations across both
    encoders, augmented only by a compact isolated rank-$r_p$
    residual per task without exemplar storage.

    \item \textbf{Flow Router, a multi-prototype conditional
    velocity-reconstruction router.} Trained via conditional flow
    matching in the full frozen CLIP feature space, the Flow Router
    scores queries against registered per-task centroids to eliminate
    the routing collapse observed in single-Gaussian routers as
    overlapping domains accumulate.

    \item \textbf{Exemplar-free functional and cross-modal
    retention with confidence gating.} EMA-anchored feature
    distillation, centroid-conditioned velocity distillation, and
    cached text-embedding alignment preserve prior knowledge across
    task boundaries, while a z-score bypass gate attenuates visual
    adapter drift on unseen domains.

\end{itemize}
\section{Related Work}
\label{sec:relatedwork}

\subsection{Continual Learning for VLMs}
Continual learning methods fall into three classical families: regularization~\cite{kirkpatrick2017overcoming, zenke2017continual, li2017learning}, memory replay~\cite{rebuffi2017icarl, lopez2017gradient, buzzega2020dark}, and architectural isolation~\cite{mallya2018packnet, yoon2017lifelong}. Applying these directly to large pre-trained VLMs risks overwriting the representations acquired during pretraining. ZSCL~\cite{zheng2023preventing} formalized this concern, introducing the MTIL benchmark and demonstrating that sequential fine-tuning rapidly erodes zero-shot transfer. CoLeCLIP~\cite{li2025coleclip} extended the problem to the open-domain class-incremental setting, where the model must classify over all accumulated classes without task identity at test time. CREST targets this same protocol.

\subsection{Parameter-Efficient Adaptation for Continual Learning}
Freezing CLIP and learning lightweight per-task modules has become standard. Visual Prompt Tuning~\cite{jia2022visual}, DualPrompt~\cite{wang2022dualprompt}, and CODA-Prompt~\cite{smith2023coda} showed that learnable tokens with task-shared and task-specific pools achieve strong performance without touching the backbone. DIKI~\cite{tang2024mind} and DPeCLIP~\cite{zhang2024dual} further reduced cross-task interference through distribution-aware calibration and prototype-conditioned prefixes. On the adapter side, LoRA~\cite{hu2021lora} and HydraLoRA~\cite{tian2024hydralora} motivated continual variants including MoE-Adapters~\cite{yu2024boosting} and C-CLIP~\cite{liu2025c}. Unlike HydraLoRA's asymmetric design, GOSA shares both projection matrices across the task sequence and confines task-specific capacity to a compact rank-$r_p$ residual, enforcing subspace orthogonality in the spirit of gradient-projection~\cite{saha2021gradient} and adapter-subspace~\cite{wang2023orthogonal} approaches. DPW~\cite{jang2026dpw} unifies prefixes and adapters with token-level gating and holds state-of-the-art results on MTIL and ODCL-CIL, but like all prior continual PEFT methods, it allocates separate parameters per task, whereas CREST keeps the dominant adapter footprint fixed.

\subsection{Task Routing under Class-Incremental Inference}
Routing a test sample to the correct per-task adapter is the central challenge of CIL. MoE-Adapters~\cite{yu2024boosting} gate on the frozen CLS token, while DIKI~\cite{tang2024mind} and DPW~\cite{jang2026dpw} select the task whose Gaussian distribution over frozen CLIP features best matches the test input. DPeCLIP~\cite{zhang2024dual} uses class-mean prototypes in place of parametric densities. As semantically adjacent fine-grained domains accumulate, frozen CLIP representations increasingly overlap and unimodal task-level densities degrade~\cite{zheng2023preventing}, increasingly failing to discriminate overlapping feature clusters. Thus, a single Gaussian per task fails to capture multimodal intra-task structure. CREST addresses this limitation through the Flow Router, which registers $K{=}8$ k-means centroids per task from the frozen CLIP feature distribution and routes by minimum conditional velocity reconstruction error computed in the full 512-dimensional frozen feature space, maintaining discrimination as tasks accumulate.

\subsection{Rectified Flow and Flow Matching}
Rectified flow~\cite{liu2022flow} learns a straight-line velocity field transporting samples from a source to a target distribution. Flow matching~\cite{lipman2022flow} generalizes this with a simulation-free regression objective over conditional probability paths. Both have been applied almost exclusively to image and video synthesis. CREST introduces the Flow Router, a prototype-conditioned 512-dimensional velocity field trained via conditional flow matching on frozen CLIP features. Crucially, the Flow Router operates directly in the full frozen feature space rather than in a compressed latent, providing both task assignment and per-sample routing confidence at inference through the conditional velocity reconstruction error to each task's stored centroids.

\subsection{Knowledge Distillation for Retention}
Learning without Forgetting~\cite{li2017learning} established that distilling a prior model's outputs during new-task training limits forgetting without exemplar storage. ZSCL~\cite{zheng2023preventing} adapted this to VLMs with feature-level distillation from a frozen backbone copy, and DPW~\cite{jang2026dpw} distills from a rolling snapshot of its prefix pool. CREST applies dual-forward functional distillation directly on the adapted feature manifold using a snapshot of the shared adapter, anchoring the full adapted representation rather than routing parameters alone. A continuous exponential moving average (EMA) of the shared adapter weights is maintained across the entire task sequence; at each task boundary, the EMA state serves as the distillation teacher, reducing drift introduced by high-loss training epochs on incoming tasks.

\subsection{Task-Vector Merging}
Task arithmetic~\cite{ilharco2022editing} showed that per-task weight deltas can be combined arithmetically to transfer behaviors between models. TIES-merging~\cite{yadav2023ties} reduced interference through magnitude trimming and sign-conflict resolution, and DARE~\cite{yu2024language} further sparsifies task vectors before summation. These methods were designed for the offline multi-task setting. Under sequential continual learning with a fixed rank budget, merging requires projecting accumulated task vectors through SVD truncation at each task boundary, discarding an increasing fraction of adapter capacity as tasks grow. Together, the above methods leave three problems unresolved: the parameter count grows with every new task, single-Gaussian routers lose discriminative power as fine-grained domains accumulate, and adapters that remain fully active progressively suppress zero-shot generalization. CREST addresses these through a shared adapter with small isolated per-task residuals, multi-prototype flow-matched routing in the full frozen feature space, and a confidence-driven bypass gate that preserves frozen CLIP transfer on unseen domains.

\section{Analysis of Limitations in Class-Incremental Learning}
\label{sec:cil-challenges}

\begin{figure}[t]
\centering
\includegraphics[width=0.95\columnwidth]{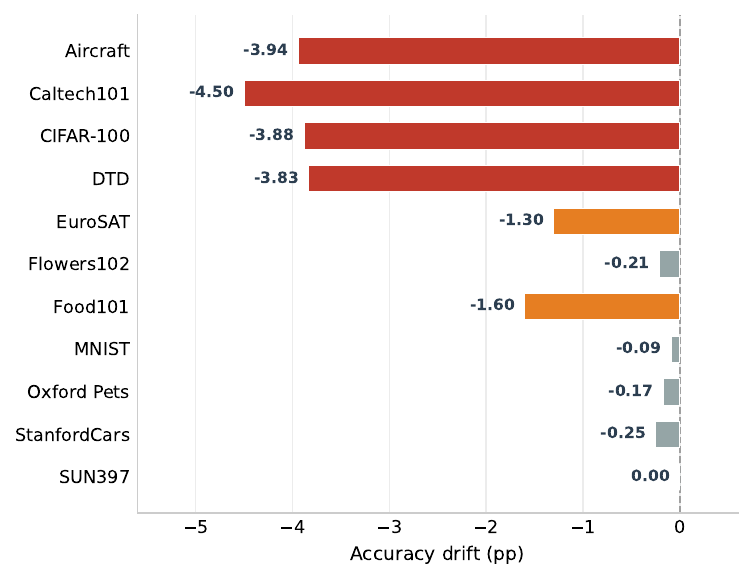}
\vspace{-0.2cm}
\caption{Per-task CIL accuracy drift (peak minus final accuracy) under single-Gaussian routing on MTIL Order-I. Tasks trained early face more competing distributions and exhibit substantially higher drift than later tasks, illustrating the routing degradation analyzed in Section~\ref{sec:cil-challenges}.}
\label{fig:cil-drift}
\vspace{-0.3cm}
\end{figure}

The limitations surveyed in the prior section become concrete failure modes under the class-incremental protocol. Under TIL evaluation, task identity is provided at inference as an oracle, so routing is never a bottleneck. CIL removes this oracle: the model must infer the task from the input alone and classify over all classes seen so far, making routing quality a primary bottleneck on final performance. Three structural limitations explain why prior methods struggle here.

\textbf{Gaussian routing degrades with task count.} Per-task Gaussians fitted on frozen CLIP features assume well-separated unimodal domains that rarely hold in practice. As semantically adjacent and fine-grained domains accumulate, the unimodal assumption increasingly fails to capture intra-task feature geometry. Tasks trained early accumulate more competing distributions and suffer the most. In an experiment on MTIL Order-I, we find that earlier tasks drift more than 3\,pp from their peak CIL accuracy, while later tasks exhibit substantially less drift (Figure~\ref{fig:cil-drift}). Caltech101 routing accuracy falls to 49.7\% under a single-Gaussian router by the end of the 11-task sequence.

\textbf{Routing errors compound classification errors.} A routing error is far more damaging than a within-task classification error. Misrouting applies a mismatched adapter and exposes the query to the wrong label set, making ground-truth recovery nearly impossible regardless of feature quality. In prior methods, the gap between TIL Last and CIL Last traces primarily to task misassignment rather than within-task difficulty.

\textbf{Always-active adapters erode zero-shot transfer and grow with task count.} Task-adapted visual modules applied to every input distort the frozen CLIP manifold that zero-shot generalization depends on, growing more disruptive as training progresses on specialized tasks. Prefix-based methods avoid this but lack consistent layer-wise adaptation for fine-grained domains, while dedicated per-task adapters solve adaptation quality at the cost of linear parameter growth. These findings motivate the core design of CREST: the Flow Router replaces single-Gaussian routing with multi-prototype conditional velocity-reconstruction scoring across $K{=}8$ k-means centroids in the full 512-dimensional frozen CLIP feature space, recovering Caltech101 routing accuracy to 95.7\% at the end of Order-I. A confidence-driven bypass gate attenuates visual adapter output on out-of-distribution inputs to preserve zero-shot transfer, and a shared GOSA core augmented by a compact per-task residual keeps dominant adapter capacity fixed. The full architecture is detailed in Section~\ref{sec:method}.

\section{Methodology}
\label{sec:method}

\begin{figure*}[t]
\centering
\includegraphics[width=\linewidth]{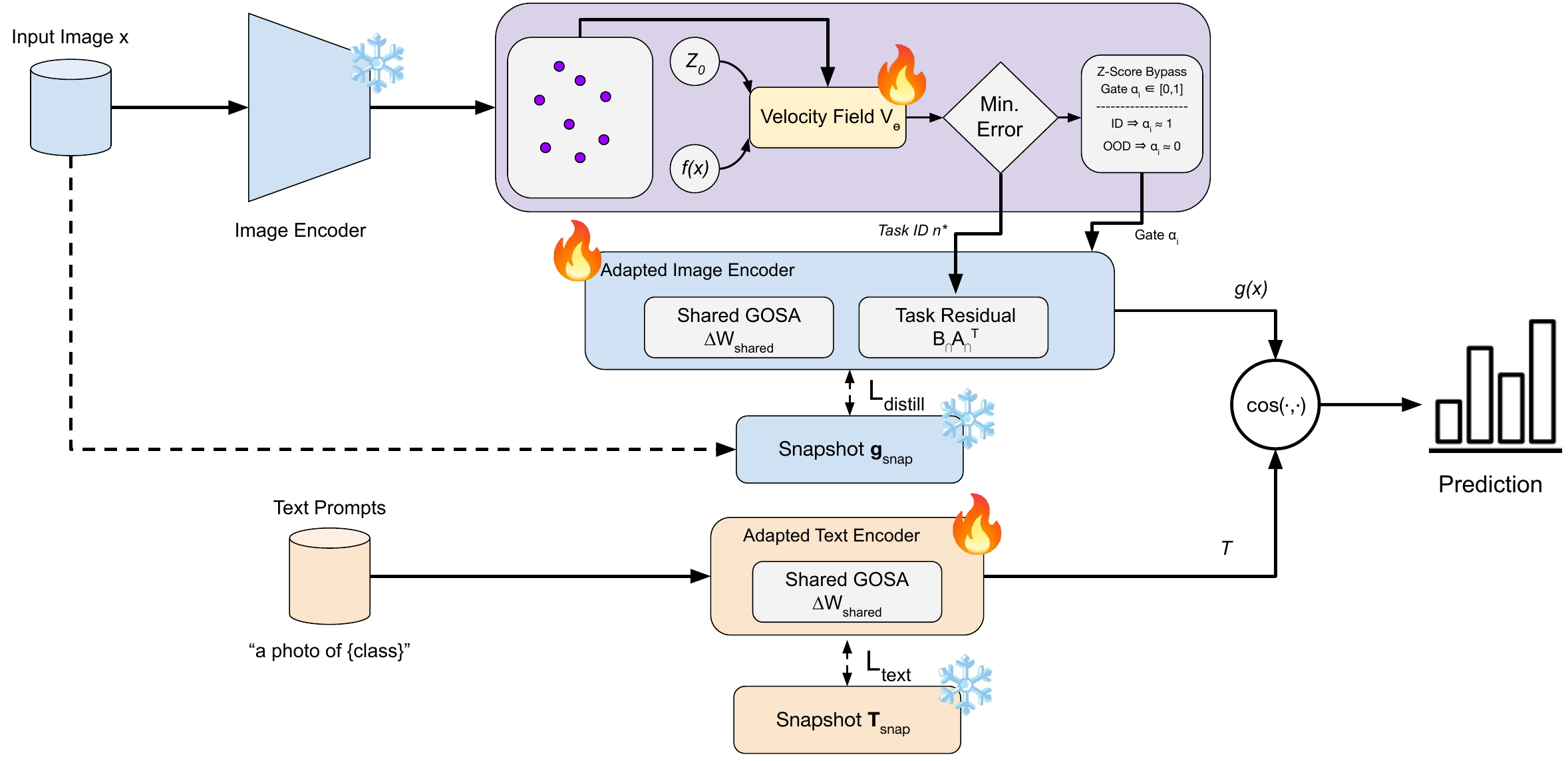}
\vspace{-0.1cm}
\caption{Overall framework of CREST. Frozen components are marked with a snowflake; trainable components with a fire icon. \textbf{GOSA} injects a shared dual-subspace delta and a per-task rank-4 residual at the QKV projection, attention output, MLP up-projection, and MLP down-projection of every transformer block in both encoders. The \textbf{Flow Router} registers $K{=}8$ k-means centroids per task from frozen CLIP features, trains a velocity field $v_\theta$ in the full 512-dimensional feature space via $\mathcal{L}_\text{CFM}$, and at CIL inference routes each query to the task whose nearest centroid yields the minimum flow reconstruction error. The \textbf{distillation arm} runs a frozen snapshot $g_\text{snap}$ alongside the live encoder $g$ to anchor adapted features without requiring exemplar storage. The \textbf{bypass gate} $\alpha_i$ scales the GOSA contribution per sample using a z-score over Flow Router transport error, recovering the frozen backbone for inputs far from any known task distribution.}
\label{fig:method}
\vspace{-0.3cm}
\end{figure*}

\subsection{Problem Formulation}
\label{sec:formulation}

We consider continual learning over $N$ sequential tasks $\{\mathcal{T}_n\}_{n=1}^{N}$. Task $\mathcal{T}_n$ provides training set $\mathcal{D}_n$ and a disjoint class set $\mathcal{C}_n$. Under the class-incremental (CIL) protocol, task identity is unavailable at inference and the model must classify over $\bigcup_{s \leq n} \mathcal{C}_s$. Under the task-incremental (TIL) protocol, task identity is provided at test time. The frozen CLIP image encoder is $f : \mathcal{X} \to \mathbb{R}^d$ with $d{=}512$; all feature vectors are $\ell_2$-normalized throughout. The GOSA-adapted encoder is $g : \mathcal{X} \to \mathbb{R}^d$. CREST keeps $f$ frozen and maintains a single shared parameter set that does not grow with $N$, augmented by a small per-task rank-$r_p$ residual. We evaluate under Order-I and Order-II of the MTIL benchmark, each presenting $N{=}11$ datasets in a different sequence.

\subsection{Gated Orthogonal Subspace Adaptation}
\label{sec:gosa}

Gated Orthogonal Subspace Adaptation (GOSA) is a shared dual-subspace adapter applied across all tasks. At each adapted weight $\mathbf{W}$, subscripts $S$ and $G$ index two disjoint rank-$r$ branches ($r{=}32$) whose down-projections $(\mathbf{W}_S, \mathbf{W}_G)$ are initialized from the top-$2r$ right singular vectors of $\mathbf{W}$ such that $\mathbf{W}_S^\top \mathbf{W}_G{=}\mathbf{0}$. Paired with shared up-projections $(\mathbf{V}_S, \mathbf{V}_G)$, they form the shared weight update
\begin{equation}
\Delta\mathbf{W}_\text{shared} = \frac{1}{r}\!\left(\mathbf{V}_S \mathbf{W}_S^\top + \mathbf{V}_G \mathbf{W}_G^\top\right).
\label{eq:delta_shared}
\end{equation}
On the vision encoder, each task $n$ also trains a rank-$r_p{=}4$ private residual $(\mathbf{A}_n, \mathbf{B}_n)$ that is frozen once task $\mathcal{T}_n$ finishes. The full per-task delta is
\begin{equation}
\Delta\mathbf{W}_n = \Delta\mathbf{W}_\text{shared} + \frac{1}{r_p}\,\mathbf{B}_n \mathbf{A}_n^\top,
\label{eq:delta_n}
\end{equation}
and the adapted layer outputs $\mathbf{h} = \mathbf{W}\mathbf{x} + \alpha \cdot \Delta\mathbf{W}_n \mathbf{x}$, where $\alpha$ is the bypass gate from Section~\ref{sec:bypass} ($\alpha{=}1$ during training). Zero-initializing the up-projections $(\mathbf{V}_S, \mathbf{V}_G, \mathbf{B}_n)$ preserves the frozen CLIP forward pass at task initialization. An orthogonality penalty
\begin{equation}
\mathcal{L}_\text{ortho} = \sum_{\text{sites}} \left\|\mathbf{W}_S^\top \mathbf{W}_G\right\|_F^2
\label{eq:ortho}
\end{equation}
keeps the two shared branches distinct during continual updates. The shared GOSA core is injected at the QKV, attention output, MLP up, and MLP down projections of every transformer block in both encoders, while per-task storage is restricted to the vision-side rank-$r_p$ residuals.

\subsection{Velocity-Field Routing}
\label{sec:flow}

\paragraph{Centroid registration.} Before training on task $\mathcal{T}_n$, $K{=}8$ k-means centroids are computed from the frozen CLIP features of $\mathcal{D}_n$ and stored in a prototype bank:
\begin{multline}
\{\mathbf{c}_{n,1}, \ldots, \mathbf{c}_{n,K}\} =
\operatorname{KMeans}\!\bigl(\{f(x) : x \in \mathcal{D}_n\},\, K\bigr), \\
\mathbf{c}_{n,j} \in \mathbb{R}^d.
\label{eq:centroids}
\end{multline}
Centroids remain fixed after registration, since $f$ is frozen throughout.

\paragraph{Velocity field training.} A shared velocity MLP $v_\theta : \mathbb{R}^d \times [0,1] \times \mathbb{R}^d \to \mathbb{R}^d$ is trained with a conditional flow-matching objective in the full $d$-dimensional frozen feature space. For each training sample on task $\mathcal{T}_n$, the nearest centroid $\mathbf{c}_{n,j}$ to $f(x)$ by cosine similarity conditions the velocity field:
\begin{align}
&\mathbf{z}_0 \sim \mathcal{N}(\mathbf{0}, \mathbf{I}_d),\quad \tau \sim \mathcal{U}[0,1],\quad \mathbf{x}_\tau = \tau\, f(x) + (1{-}\tau)\,\mathbf{z}_0, \nonumber \\
&\mathcal{L}_\text{CFM} = \left\|v_\theta\!\left(\mathbf{x}_\tau,\,\tau,\,\mathbf{c}_{n,j}\right) - \bigl(f(x) - \mathbf{z}_0\bigr)\right\|^2.
\label{eq:cfm}
\end{align}
Training on frozen features $f(x)$ ensures that the feature distribution evaluated by $v_\theta$ at inference matches its training distribution across the entire task sequence. Operating directly in the full 512-dimensional space avoids both ill-conditioned covariance estimation in high dimensions and the discriminative information loss of compressed latent bottlenecks.

\paragraph{Routing via conditional velocity reconstruction error.}
Rather than integrating a probability-flow ODE likelihood, CREST scores prototypes at CIL inference via a conditional velocity reconstruction residual evaluated across $M$ uniformly spaced interpolation levels with shared noise
$\mathbf{z}_0 \sim \mathcal{N}(\mathbf{0}, \mathbf{I}_d)$:
\begin{equation}
\begin{split}
e(x, \mathbf{c}) = \frac{1}{M} \sum_{m=1}^{M}
\left\|v_\theta\!\left(\mathbf{x}_{\tau_m},\, \tau_m,\,
\mathbf{c}\right) - \bigl(f(x) - \mathbf{z}_0\bigr)\right\|^2, \\
\mathbf{x}_{\tau_m} = \tau_m\, f(x) + (1{-}\tau_m)\,\mathbf{z}_0.
\end{split}
\label{eq:terror}
\end{equation}
Because the regression target $f(x) - \mathbf{z}_0$ is independent of $\mathbf{c}$, $e(x, \mathbf{c})$ acts as a learned conditional energy score rather than an ODE transport distance: a centroid trained far from $x$ produces a large residual, providing a discrimination signal for task assignment. Multi-scale evaluation across $\tau_m \in [0,1]$, together with the margin objective (Eq.~\ref{eq:total}), enables $v_\theta$ to model local manifold geometry per prototype without explicit covariance estimation.
The query is assigned to the task minimising this error:
\begin{equation}
n^* = \underset{n \leq n_\text{learnt},\; j \in [K]}
{\operatorname{arg\,min}}\; e(x, \mathbf{c}_{n,j}).
\label{eq:route}
\end{equation}
This routing pass operates on frozen features $f(x)$, requiring no additional backbone forward pass.

\begin{table*}[t]
\centering
\caption{Per-dataset comparison on MTIL Order-I (TIL protocol), reporting Transfer, Avg, and Last (\%). Best results are \textbf{bold}, second-best \underline{underlined}. $\dagger$ denotes methods requiring auxiliary data. Baseline values from~\cite{jang2026dpw}.}
\label{tab:mtil-o1}
\resizebox{\textwidth}{!}{%
\begin{tabular}{l|ccccccccccc|c}
\hline
Method & Aircraft & Caltech101 & CIFAR100 & DTD & EuroSAT & Flowers & Food & MNIST & OxfordPet & Cars & SUN397 & Average \\
\hline
\multicolumn{13}{l}{\textit{Transfer}} \\
ZSCL~\cite{zheng2023preventing}          & -- & 86.0 & 67.4 & 45.4 & 50.4 & 69.1 & \underline{87.6} & 61.8 & 86.8 & 60.1 & 66.8 & 68.1 \\
DIKI~\cite{tang2024mind}                 & -- & 92.9 & 69.0 & 43.2 & 48.2 & 67.4 & 85.2 & 63.0 & 87.9 & 63.8 & 66.3 & 68.7 \\
MoE-Adapters~\cite{yu2024boosting}       & -- & 87.9 & 68.2 & 44.4 & 49.9 & 70.7 & \textbf{88.7} & 59.7 & 89.1 & 64.5 & 65.5 & 68.9 \\
GIFT$^\dagger$~\cite{wu2025synthetic}    & -- & 88.5 & \underline{69.8} & \underline{46.0} & 49.4 & 68.5 & 87.1 & \textbf{69.9} & 88.9 & 57.7 & 67.7 & 69.3 \\
DPW~\cite{jang2026dpw}             & -- & \underline{92.9} & 69.0 & 45.3 & \underline{54.2} & \underline{71.1} & 86.2 & 63.8 & 89.3 & \underline{65.5} & 66.5 & \underline{70.4} \\
\rowcolor{gray!20}
CREST (Ours) & -- & \textbf{95.4} & \textbf{71.6} & \textbf{47.9} & \textbf{57.7} & \textbf{75.5} & \underline{88.0} & \underline{67.1} & \textbf{90.4} & \textbf{68.0} & \textbf{68.7} & \textbf{73.1 \textcolor{red}{(+2.7\%)}} \\
\hline
\multicolumn{13}{l}{\textit{Average}} \\
ZSCL~\cite{zheng2023preventing}          & 45.1 & 92.0 & 80.1 & 64.3 & 79.5 & 81.6 & \underline{89.6} & 75.2 & 88.9 & 64.7 & 68.0 & 75.4 \\
DIKI~\cite{tang2024mind}                 & 45.1 & 95.5 & 83.1 & 64.8 & 79.9 & 83.5 & 87.0 & 76.2 & 89.6 & 67.0 & 67.1 & 76.3 \\
MoE-Adapters~\cite{yu2024boosting}       & 50.2 & 91.9 & 83.1 & 69.4 & 78.9 & 84.0 & 89.1 & 73.7 & 89.3 & 67.7 & 66.9 & 76.7 \\
GIFT$^\dagger$~\cite{wu2025synthetic}    & 51.9 & 93.9 & 81.4 & 67.7 & \underline{80.3} & 82.8 & 89.3 & \textbf{80.6} & 90.3 & 63.1 & \textbf{68.9} & 77.3 \\
DPW~\cite{jang2026dpw}             & \underline{60.4} & \underline{96.2} & \underline{84.5} & \underline{70.8} & \textbf{82.4} & \underline{85.8} & 87.7 & \underline{76.8} & \underline{90.9} & \textbf{69.7} & 67.6 & \underline{79.3} \\
\rowcolor{gray!20}
CREST (Ours) & \textbf{65.9} & \textbf{96.6} & \textbf{86.8} & \textbf{76.2} & 79.5 & \textbf{86.1} & \textbf{89.7} & 71.5 & \textbf{91.6} & \underline{69.5} & \textbf{69.8} & \textbf{80.3 \textcolor{red}{(+1.0\%)}} \\
\hline
\multicolumn{13}{l}{\textit{Last}} \\
ZSCL~\cite{zheng2023preventing}          & 40.6 & 92.2 & 81.3 & 70.5 & 94.8 & 90.5 & \underline{91.9} & 98.7 & 93.9 & 85.3 & \underline{80.2} & 83.6 \\
DIKI~\cite{tang2024mind}                 & 45.2 & 95.7 & 86.3 & 72.9 & 98.0 & 97.0 & 89.2 & 99.4 & 94.2 & 81.6 & 76.6 & 85.1 \\
MoE-Adapters~\cite{yu2024boosting}       & 49.8 & 92.2 & 86.1 & 78.1 & 95.7 & 94.3 & 89.5 & 98.1 & 89.9 & 81.6 & 80.0 & 85.0 \\
GIFT$^\dagger$~\cite{wu2025synthetic}    & 47.9 & 95.6 & 82.8 & 75.1 & 97.3 & 94.2 & 91.7 & 99.2 & 94.2 & \underline{87.0} & \textbf{80.9} & 86.0 \\
DPW~\cite{jang2026dpw}             & \underline{60.5} & \underline{96.6} & \underline{87.9} & \underline{80.4} & \textbf{98.5} & \underline{98.1} & 89.4 & \textbf{99.6} & \underline{95.1} & 86.7 & 78.8 & \underline{88.3} \\
\rowcolor{gray!20}
CREST (Ours) & \textbf{64.5} & \underline{96.6} & \textbf{90.1} & \textbf{87.3} & \underline{98.2} & \textbf{98.4} & \textbf{93.5} & 99.2 & \textbf{95.9} & \textbf{87.4} & 80.0 & \textbf{90.1 \textcolor{red}{(+1.8\%)}} \\
\hline
\end{tabular}}
\end{table*}

\begin{table*}[t]
\centering
\caption{Per-dataset comparison on MTIL Order-II (TIL protocol), reporting Transfer, Avg, and Last (\%). Best results are \textbf{bold}, second-best \underline{underlined}. $\dagger$ denotes methods requiring auxiliary data.}
\label{tab:order2}
\resizebox{\textwidth}{!}{%
\begin{tabular}{l|ccccccccccc|c}
\hline
Method & Cars & Food & MNIST & OxfordPet & Flowers & SUN397 & Aircraft & Caltech101 & DTD & EuroSAT & CIFAR100 & Average \\
\hline
\multicolumn{13}{l}{\textit{Transfer}} \\
ZSCL~\cite{zheng2023preventing}          & -- & 88.3 & 57.5 & 84.7 & 68.1 & 64.8 & 21.1 & 88.2 & 45.3 & \textbf{55.2} & 68.2 & 64.2 \\
DIKI~\cite{tang2024mind}                 & -- & 85.8 & 59.8 & 89.1 & 71.8 & 62.6 & 24.3 & 93.3 & 42.7 & 46.8 & 67.8 & 64.4 \\
MoE-Adapters~\cite{yu2024boosting}       & -- & \textbf{88.8} & 59.5 & \underline{89.1} & 69.9 & 64.4 & 18.1 & 86.9 & 43.7 & 54.6 & 68.2 & 64.3 \\
GIFT$^\dagger$~\cite{wu2025synthetic}    & -- & 88.3 & 63.4 & 88.1 & 70.8 & \textbf{67.7} & 22.8 & 90.4 & \textbf{46.7} & \underline{51.8} & 68.8 & \underline{65.9} \\
DPW~\cite{jang2026dpw}             & -- & 85.9 & \underline{63.3} & \textbf{89.4} & \underline{72.2} & 64.2 & \underline{24.6} & \underline{94.0} & 44.6 & 49.4 & \underline{69.1} & 65.7 \\
\rowcolor{gray!20}
CREST (Ours)                             & -- & 87.4 & \textbf{66.2} & \textbf{89.7} & \textbf{74.4} & \underline{66.2} & \textbf{26.0} & \textbf{94.9} & \textbf{46.7} & 50.6 & \textbf{71.0} & \textbf{67.3 \textcolor{red}{(+1.6\%)}} \\
\hline
\multicolumn{13}{l}{\textit{Average}} \\
ZSCL~\cite{zheng2023preventing}          & 81.7 & \textbf{91.3} & 91.0 & 91.0 & 82.9 & 72.5 & 33.6 & 89.7 & 53.3 & \textbf{62.8} & 69.9 & 74.5 \\
DIKI~\cite{tang2024mind}                 & 81.9 & 88.9 & 92.1 & 92.8 & 87.7 & 70.3 & 34.3 & 94.2 & 51.5 & 56.1 & 69.5 & 74.5 \\
MoE-Adapters~\cite{yu2024boosting}       & 84.9 & 89.9 & 89.3 & 91.4 & 86.2 & 72.2 & 33.4 & 89.4 & 53.3 & 61.4 & 69.9 & 74.7 \\
GIFT$^\dagger$~\cite{wu2025synthetic}    & 83.2 & 90.8 & 92.6 & 92.8 & 85.8 & \textbf{74.1} & 36.0 & 92.1 & 54.7 & 60.0 & 70.4 & 75.7 \\
DPW~\cite{jang2026dpw}             & \underline{86.6} & 88.9 & \underline{92.8} & \underline{93.6} & \underline{88.5} & 71.5 & \underline{40.4} & \underline{94.9} & \underline{54.2} & \underline{58.3} & \underline{70.9} & \underline{76.4} \\
\rowcolor{gray!20}
CREST (Ours)                             & \textbf{87.8} & \textbf{92.9} & \textbf{93.4} & \textbf{94.3} & \textbf{89.7} & \underline{73.7} & \textbf{44.0} & \textbf{95.5} & \textbf{58.0} & \textbf{59.3} & \textbf{72.1} & \textbf{78.2 \textcolor{red}{(+1.8\%)}} \\
\hline
\multicolumn{13}{l}{\textit{Last}} \\
ZSCL~\cite{zheng2023preventing}          & 78.2 & \underline{91.1} & 97.6 & 92.5 & 87.4 & 78.2 & 45.0 & 92.3 & 72.7 & 96.2 & 86.3 & 83.4 \\
DIKI~\cite{tang2024mind}                 & 81.9 & 89.2 & \underline{99.4} & 94.3 & 96.8 & 76.7 & 46.3 & 95.9 & 74.8 & 98.3 & 86.6 & 85.5 \\
MoE-Adapters~\cite{yu2024boosting}       & 84.1 & 88.5 & 94.0 & 91.8 & 94.1 & 77.8 & 50.4 & 93.3 & 77.1 & 87.7 & 86.6 & 84.1 \\
GIFT$^\dagger$~\cite{wu2025synthetic}    & 81.0 & 90.2 & 98.6 & 94.0 & 91.5 & 78.6 & 51.7 & 94.6 & 75.6 & 95.4 & 86.6 & 85.3 \\
DPW~\cite{jang2026dpw}             & \underline{86.6} & 89.2 & \textbf{99.6} & \underline{95.1} & \underline{97.9} & \underline{78.4} & \underline{59.4} & \textbf{96.4} & \underline{79.8} & \textbf{98.4} & \textbf{87.9} & \underline{88.1} \\
\rowcolor{gray!20}
CREST (Ours)                             & \textbf{87.8} & \textbf{93.4} & \underline{99.4} & \textbf{96.1} & \textbf{98.4} & \textbf{79.9} & \textbf{65.5} & \underline{94.9} & \textbf{88.0} & \underline{98.3} & 84.3 & \textbf{89.7 \textcolor{red}{(+1.6\%)}} \\
\hline
\end{tabular}}
\end{table*}

\subsection{Functional Distillation}
\label{sec:distill}

To stabilize the shared subspace across task boundaries, CREST
maintains an EMA of the shared GOSA parameters
($\bar{\theta}_\text{shared} \leftarrow \eta\,\bar{\theta}_\text{shared}
+ (1{-}\eta)\,\theta_\text{shared}$, $\eta{=}0.999$) and freezes this
smoothed state into a snapshot encoder $g_\text{snap}$ and velocity
snapshot $v_\text{snap}$ at each task boundary. The feature-level
distillation loss anchors the live encoder to its prior manifold:
\begin{equation}
\mathcal{L}_\text{distill} = 1 - \frac{1}{B} \sum_{i=1}^{B}
\cos\!\left(g(x_i),\; g_\text{snap}(x_i)\right),
\label{eq:distill}
\end{equation}
where $\cos(\cdot,\cdot)$ denotes cosine similarity.
To prevent $v_\theta$ from drifting on prior routing directions, a
velocity-snapshot loss anchors it to $v_\text{snap}$ over noise
conditioned on all prior centroids:
\begin{equation}
\mathcal{L}_\text{flow} = \frac{1}{n} \sum_{k < n} \sum_{j=1}^{K}
\mathbb{E}_{\mathbf{z},\, \tau} \left\|v_\theta(\mathbf{z}, \tau,
\mathbf{c}_{k,j}) - v_\text{snap}(\mathbf{z}, \tau,
\mathbf{c}_{k,j})\right\|^2.
\label{eq:flowlwf}
\end{equation}
No image exemplars are required. Rather, frozen centroids
$\{\mathbf{c}_{k,j}\}_{k < n}$ serve as the sole conditioning signal.
Text-encoder drift is suppressed via cached snapshot features
$\mathbf{T}^\text{snap}_{k,c}$:
\begin{equation}
\mathcal{L}_\text{text} = \frac{1}{n} \sum_{k < n} \frac{1}{|\mathcal{C}_k|}
\sum_{c \in \mathcal{C}_k} \left(1 - \cos\!\left(\mathbf{T}^\text{live}_{k,c},\;
\mathbf{T}^\text{snap}_{k,c}\right)\right),
\label{eq:textdistill}
\end{equation}
where $\mathbf{T}^\text{live}_{k,c}$ is the live text encoder output
for class $c$ of task $k$.

\subsection{Residual Bypass Gate}
\label{sec:bypass}

The routing pass of Eq.~\ref{eq:route} returns the per-sample minimum transport error $d_{\min,i} = \min_{n,j}\, e(x_i, \mathbf{c}_{n,j})$ at no additional cost. After training each task $\mathcal{T}_n$, a brief calibration pass over $\mathcal{D}_n$ records the empirical mean $\mu_n$ and standard deviation $s_n$ of in-distribution reconstruction errors. The per-sample bypass gate is then
\begin{equation}
\alpha_i = \sigma\!\left(\frac{\beta\,\bigl(\mu_{\hat{n}_i} + \gamma\, s_{\hat{n}_i} - d_{\min,i}\bigr)}{s_{\hat{n}_i} + \epsilon}\right), \quad \beta{=}3.0,\quad \gamma{=}2.0,
\label{eq:alpha}
\end{equation}
where $\sigma$ is the sigmoid. Here, $\beta{=}3.0$ controls the sigmoid sharpness, and $\gamma{=}2.0$ places the soft threshold two standard deviations above the in-distribution mean, requiring only two stored scalars per task. An in-distribution input yields small $d_{\min,i}$, giving $\alpha_i \approx 1$ and full visual GOSA activation. An out-of-distribution input drives $\alpha_i \to 0$, reverting the vision encoder to frozen CLIP while text GOSA remains active to preserve cross-modal alignment.

\subsection{Training Objective}
\label{sec:training}

CREST trains each task $\mathcal{T}_n$ with the combined loss
\begin{multline}
\mathcal{L} = \mathcal{L}_\text{CE} + \mathcal{L}_\text{CFM} + \lambda_o\,\mathcal{L}_\text{ortho} \\
+ \mathbbm{1}[n{>}0]\!\bigl(\lambda_d\,\mathcal{L}_\text{distill} + \lambda_f\,\mathcal{L}_\text{flow} \\
+ \lambda_m\,\mathcal{L}_\text{margin} + \lambda_\text{td}\,\mathcal{L}_\text{text}\bigr),
\label{eq:total}
\end{multline}
where $\mathcal{L}_\text{margin}$ is a contrastive margin loss on $v_\theta$ that
pushes current-task transport errors below all prior tasks by a margin
$\delta{=}0.1$. 
\section{Experiments}
\label{sec:experiments}

\subsection{Experimental Setup}

\begin{table*}[t]
\centering
\caption{Per-dataset comparison on ODCL-CIL (Order-I), reporting Transfer, Avg, and Last (\%). Best results are \textbf{bold}, second-best \underline{underlined}. Baseline values are reported from~\cite{jang2026dpw}.}
\label{tab:odcl-cil}
\resizebox{\textwidth}{!}{%
\begin{tabular}{l|ccccccccccc|c}
\hline
Method & Aircraft & Caltech101 & CIFAR100 & DTD & EuroSAT & Flowers & Food & MNIST & OxfordPet & Cars & SUN397 & Average \\
\hline
\multicolumn{13}{l}{\textit{Transfer}} \\
ZSCL~\cite{zheng2023preventing}    & -- & 84.6 & 67.5 & 44.8 & 51.5 & 69.0 & 87.6 & 62.3 & 87.1 & 59.7 & 66.4 & 68.0 \\
MoE-Adapters~\cite{yu2024boosting} & -- & 88.2 & 66.8 & 44.7 & 54.1 & 70.6 & \underline{88.4} & 59.5 & 89.0 & 64.7 & 65.0 & 69.1 \\
CoLeCLIP~\cite{li2025coleclip}     & -- & 88.2 & 65.1 & 44.7 & 54.1 & 68.8 & \textbf{88.5} & 59.5 & 89.0 & 64.7 & 65.1 & 68.8 \\
DPeCLIP~\cite{zhang2024dual}       & -- & 88.2 & 67.2 & 44.7 & 54.0 & 70.6 & 88.2 & 59.5 & 89.0 & 64.7 & 64.8 & 69.1 \\
DPW~\cite{jang2026dpw}             & -- & \underline{92.9} & \underline{69.0} & \underline{45.3} & \underline{54.2} & \underline{71.1} & 86.2 & \underline{63.8} & \underline{89.3} & \underline{65.7} & \underline{66.5} & \underline{70.4} \\
\rowcolor{gray!20}
CREST (Ours)                       & -- & \textbf{94.3} & \textbf{76.7} & \textbf{48.1} & \textbf{55.8} & \textbf{73.5} & 87.8 & \textbf{65.3} & \textbf{91.2} & \textbf{67.3} & \textbf{68.5} & \textbf{72.8 \textcolor{red}{(+2.4\%)}} \\
\hline
\multicolumn{13}{l}{\textit{Average}} \\
ZSCL~\cite{zheng2023preventing}    & 46.3 & 68.3 & 74.3 & 56.3 & 79.1 & 81.4 & 84.9 & \underline{74.0} & 89.0 & 64.4 & \underline{67.5} & 71.8 \\
MoE-Adapters~\cite{yu2024boosting} & 37.2 & 65.3 & 79.5 & 67.6 & 19.7 & 83.1 & 80.5 & \underline{74.0} & 88.5 & 67.5 & 65.3 & 66.2 \\
CoLeCLIP~\cite{li2025coleclip}     & 48.2 & 77.8 & 71.7 & 65.7 & 76.8 & 83.8 & \textbf{89.6} & 72.2 & 90.3 & 68.0 & 66.4 & 73.7 \\
DPeCLIP~\cite{zhang2024dual}       & 49.9 & 85.3 & 81.5 & 65.3 & 81.6 & 84.3 & 88.9 & \underline{74.0} & \underline{90.4} & 68.3 & 66.2 & 76.1 \\
DPW~\cite{jang2026dpw}             & \underline{60.2} & \underline{93.5} & \underline{83.9} & \underline{68.9} & \underline{82.0} & \underline{84.8} & 87.6 & \textbf{76.7} & \underline{90.4} & \underline{69.4} & \textbf{67.6} & \underline{78.6} \\
\rowcolor{gray!20}
CREST (Ours)                       & \textbf{64.8} & \textbf{96.1} & \textbf{91.5} & \textbf{74.7} & \textbf{82.3} & \textbf{86.0} & \underline{89.4} & 73.3 & \textbf{91.2} & \textbf{69.9} & 66.2 & \textbf{80.5 \textcolor{red}{(+1.9\%)}} \\
\hline
\multicolumn{13}{l}{\textit{Last}} \\
ZSCL~\cite{zheng2023preventing}    & 42.5 & 64.4 & 67.2 & 54.8 & 89.7 & 90.4 & \underline{91.7} & 95.8 & 93.4 & 85.2 & 78.3 & 77.6 \\
MoE-Adapters~\cite{yu2024boosting} & 34.1 & 47.6 & 80.9 & 75.5 & 0.0  & 93.0 & 70.8 & \textbf{99.4} & 86.4 & 79.8 & 68.9 & 66.9 \\
CoLeCLIP~\cite{li2025coleclip}     & 48.1 & 73.1 & 65.2 & 69.6 & 84.0 & \underline{96.2} & 90.9 & 94.6 & 93.5 & 82.6 & 79.3 & 79.7 \\
DPeCLIP~\cite{zhang2024dual}       & 49.9 & 84.2 & 83.2 & 71.1 & 97.0 & 95.8 & \textbf{92.0} & \textbf{99.4} & \underline{93.9} & 84.5 & \underline{80.2} & 84.6 \\
DPW~\cite{jang2026dpw}             & \underline{60.1} & \underline{91.1} & \underline{85.9} & \underline{75.7} & \textbf{97.9} & \underline{96.2} & 89.1 & \underline{99.2} & 93.0 & \underline{86.0} & 78.3 & \underline{86.6} \\
\rowcolor{gray!20}
CREST (Ours)                       & \textbf{63.1} & \textbf{95.6} & \textbf{93.5} & \textbf{82.8} & \underline{97.3} & \textbf{97.0} & 91.6 & \underline{99.2} & \textbf{94.7} & \textbf{88.6} & \textbf{81.3} & \textbf{89.5 \textcolor{red}{(+2.9\%)}} \\
\hline
\end{tabular}}
\end{table*}

\textbf{Benchmark.} We evaluate on the MTIL benchmark~\cite{zheng2023preventing}, comprising 11 datasets spanning Aircraft \cite{maji2013fine}, Caltech101 \cite{FeiFei2004LearningGV}, CIFAR-100 \cite{Krizhevsky09learningmultiple}, DTD \cite{cimpoi14describing}, EuroSAT \cite{helber2017eurosat}, Flowers102 \cite{Nilsback08}, Food101 \cite{bossard14}, MNIST \cite{lecun2010mnist}, Oxford Pets \cite{parkhi12a}, StanfordCars \cite{6755945}, and SUN397 \cite{Xiao:2010} (1,201 classes total), under two task orderings and both TIL and ODCL-CIL protocols. We report \textit{Transfer} (accuracy after the first task, measuring zero-shot retention), \textit{Last} (accuracy after all tasks), and \textit{Avg} (mean over the full task-by-task matrix).

\noindent \textbf{Baselines.} We compare against Zero-shot CLIP, ZSCL~\cite{zheng2023preventing}, DIKI~\cite{tang2024mind}, MoE-Adapters~\cite{yu2024boosting}, GIFT~\cite{wu2025synthetic}, CoLeCLIP~\cite{li2025coleclip}, DPeCLIP~\cite{zhang2024dual}, and DPW~\cite{jang2026dpw}. Methods marked $\dagger$ rely on auxiliary data not available at test time. All baselines either grow parameters with the number of tasks or require exemplar storage, whereas CREST requires neither.

\noindent \textbf{Implementation Details.} All experiments use a frozen ViT-B/16 CLIP backbone. GOSA uses shared rank $r{=}32$ with per-task rank $r_p{=}4$. All backbone weights are frozen throughout training; GOSA parameters are updated with SGD (peak learning rate $0.05$, cosine schedule, one warmup epoch, batch size 128) and the velocity MLP $v_\theta$ with AdamW (learning rate $3{\times}10^{-4}$, full precision). The loss weights are set to $\lambda_d{=}10.0$, $\lambda_f{=}1.5$, $\lambda_m{=}0.25$, $\lambda_\text{td}{=}5.0$, $\lambda_o{=}0.01$. Each task trains for up to 12 epochs with early stopping (patience\,=\,3). Text embeddings use \texttt{a photo of \{class\}}. All results are averaged over three seeds.

\subsection{Main Results}

\subsubsection{Task-Incremental Learning}

Tables~\ref{tab:mtil-o1} and~\ref{tab:order2} report per-dataset TIL results on Order-I and Order-II. CREST achieves 73.1\% Transfer, 80.3\% Avg, and 90.1\% Last on Order-I (+2.7, +1.0, and +1.8\,pp over DPW), and 67.3\% Transfer, 78.2\% Avg, and 89.7\% Last on Order-II (+1.6, +1.8, and +1.6\,pp). Gains hold across both orderings and are strongest on fine-grained Aircraft and DTD, while EuroSAT and MNIST near saturation across all methods. A shared GOSA core with compact rank-$r_p$ residuals and EMA-anchored distillation outperforms full per-task $O(N)$ adapters without linear parameter growth.

\subsubsection{Class-Incremental Learning}

Table~\ref{tab:odcl-cil} reports ODCL-CIL results under Order-I. Removing the task-identity oracle requires the Flow Router to infer the task from the query image to select the active per-task residual while evaluating logits over the union of all seen classes. CREST achieves 72.8\% Transfer, 80.5\% Avg, and 89.5\% Last, surpassing DPW by 2.4, 1.9, and 2.9\,pp. The average TIL-to-CIL Last gap is 0.6\,pp (90.1\% vs.\ 89.5\%), indicating that the velocity-field router maintains high task assignment accuracy (95.4\%; Table~\ref{tab:router-ablation}) across the full 11-task sequence. DTD shows the largest per-dataset drop from TIL to CIL (87.3\% vs.\ 82.8\%), consistent with its fine-grained texture features overlapping with other domains in CLIP space. MNIST and EuroSAT are near-perfect under both protocols, reflecting their visually distinct distributions. The substantial CIL Transfer gain over DPW (+2.4\,pp) shows that the bypass gate effectively attenuates visual adapter drift on unseen domains when routing confidence is low.

\subsection{Ablation Study}
\label{sec:ablation}

To isolate each module's contribution under identical GOSA weights and frozen CLIP features ($d{=}512$), Tables~\ref{tab:router-ablation} and~\ref{tab:component-ablation} report controlled ablations on MTIL Order-I (ODCL-CIL) across three seeds.

\begin{table}[t]
\centering
\caption{Matched-control router comparison on MTIL Order-I (ODCL-CIL) using the same trained GOSA checkpoint and frozen 512-d CLIP features (mean $\pm$ std over 3 seeds).}
\label{tab:router-ablation}
\setlength{\tabcolsep}{3.8pt}
\renewcommand{\arraystretch}{0.95}
\resizebox{\columnwidth}{!}{%
\begin{tabular}{l|c|cccc}
\hline
Router Variant & $K$ & Route Acc. (\%) & CIL Avg (\%) & CIL Last (\%) & Latency (ms) \\
\hline
Single-Gaussian (512-d)      & 1  & 64.2 $\pm$ 0.8 & 76.4 $\pm$ 0.4 & 84.8 $\pm$ 0.5 & 0.3 \\
Nearest-Centroid (Cosine)    & 8  & 79.4 $\pm$ 0.5 & 78.8 $\pm$ 0.3 & 87.3 $\pm$ 0.3 & 0.2 \\
Full-Covariance GMM (512-d)  & 8  & 83.1 $\pm$ 0.6 & 79.2 $\pm$ 0.3 & 87.9 $\pm$ 0.4 & 0.8 \\
Learned MLP Router (512-d)   & -- & 86.5 $\pm$ 0.7 & 79.4 $\pm$ 0.4 & 88.1 $\pm$ 0.4 & 0.3 \\
\rowcolor{gray!20}
\textbf{CREST Flow Router}   & 8  & \textbf{95.4 $\pm$ 0.3} & \textbf{80.5 $\pm$ 0.2} & \textbf{89.5 $\pm$ 0.2} & 1.1 \\
\hline
Oracle Task ID (CIL)         & -- & 100.0 & 80.9 $\pm$ 0.2 & 89.9 $\pm$ 0.2 & -- \\
\hline
\end{tabular}}
\vspace{-0.2cm}
\end{table}

\begin{table}[t]
\centering
\caption{Component ablation and total persistent parameter count at $N{=}11$ tasks on MTIL Order-I (ODCL-CIL, mean $\pm$ std over 3 seeds).}
\label{tab:component-ablation}
\setlength{\tabcolsep}{4.5pt}
\renewcommand{\arraystretch}{0.95}
\resizebox{\columnwidth}{!}{%
\begin{tabular}{l|ccc|c}
\hline
Configuration & Transfer (\%) & CIL Avg (\%) & CIL Last (\%) & Params (M) \\
\hline
\rowcolor{gray!20}
\textbf{Full CREST}                        & \textbf{72.8 $\pm$ 0.3} & \textbf{80.5 $\pm$ 0.2} & \textbf{89.5 $\pm$ 0.2} & 23.32 \\
Rank-64 LoRA (replace GOSA)                & 71.4 $\pm$ 0.4 & 79.1 $\pm$ 0.3 & 87.9 $\pm$ 0.3 & 23.32 \\
w/o Private Residuals ($r_p{=}0$)          & 72.6 $\pm$ 0.3 & 78.4 $\pm$ 0.3 & 87.2 $\pm$ 0.3 & 16.83 \\
w/o Bypass Gate ($\alpha{=}1$)             & 69.1 $\pm$ 0.4 & 79.2 $\pm$ 0.2 & 89.4 $\pm$ 0.2 & 23.32 \\
w/o EMA Teacher (Hard Snapshot)            & 72.1 $\pm$ 0.4 & 79.6 $\pm$ 0.3 & 88.4 $\pm$ 0.3 & 23.32 \\
w/o Feature Distill ($\lambda_d{=}0$)      & 71.8 $\pm$ 0.4 & 78.7 $\pm$ 0.3 & 87.5 $\pm$ 0.4 & 23.32 \\
w/o Velocity Distill ($\lambda_f{=}0$)     & 72.4 $\pm$ 0.4 & 78.9 $\pm$ 0.4 & 87.4 $\pm$ 0.4 & 23.32 \\
w/o Text Distill ($\lambda_\text{td}{=}0$) & 70.2 $\pm$ 0.5 & 77.8 $\pm$ 0.4 & 86.8 $\pm$ 0.4 & 23.32 \\
\hline
\end{tabular}}
\vspace{-0.2cm}
\end{table}

\paragraph{Key observations.}
\textbf{(1) Router design (Table~\ref{tab:router-ablation}).}
On identical frozen 512-d CLIP features, the Flow Router outperforms $K{=}8$ full-covariance GMM and a sequentially trained MLP router by $+12.3$\,pp and $+8.9$\,pp in routing accuracy ($+1.6$\,pp and $+1.4$\,pp in CIL Last) with only $0.9$\,ms added latency over cosine lookup, confirming that multi-scale velocity residuals capture local manifold geometry without router forgetting. Reaching $95.4\%$ routing accuracy brings CIL Last within $0.4$\,pp of the Oracle-Routed CIL upper bound ($89.9\%$) and within $0.6$\,pp of the TIL Last reference ($90.1\%$).

\textbf{(2) Architecture and retention (Table~\ref{tab:component-ablation}).}
GOSA's orthogonal dual-subspace sharing outperforms a parameter-matched rank-64 LoRA by $+1.6$\,pp CIL Last, while compact vision-only $r_p{=}4$ private residuals add $+2.3$\,pp for only $0.59$\,M parameters per task ($16.83$\,M $\to$ $23.32$\,M at $N{=}11$). Disabling the bypass gate ($\alpha{=}1$) leaves CIL Last intact ($89.4\%$) but drops Transfer by $3.7$\,pp due to unattenuated visual drift on unseen domains. EMA smoothing adds $+1.1$\,pp, and the three distillation terms—led by text-embedding anchoring ($+2.7$\,pp), which stabilizes the always-active text adapter—curb cross-task forgetting. Sensitivity to $(\beta, \gamma, \delta, M)$ and noise seed $\mathbf{z}_0$ is reported in the supplemental material.
\section{Conclusion}
\label{sec:conclusion}

CREST mitigates linear parameter growth, routing degradation, and zero-shot erosion by pairing a shared GOSA core and compact per-task residuals with a 512-dimensional flow-matching router and confidence-driven bypass gate. On MTIL, CREST surpasses prior state of the art across all three settings, achieving 73.1\% Transfer, 80.3\% Avg, and 90.1\% Last on TIL Order-I, 67.3\%, 78.2\%, and 89.7\% on TIL Order-II, and 72.8\%, 80.5\%, and 89.5\% on ODCL-CIL, with a 0.6\,pp TIL-to-CIL Last gap confirming reliable multi-prototype routing across 11 tasks.

\section*{LLM Usage Statement}
A large language model (LLM) was used to assist with editing and refining the manuscript text to improve readability, clarity, and understandability. 
{
    \small
    \bibliographystyle{ieeenat_fullname}
    \bibliography{main}
}


\end{document}